\documentclass[letterpaper, 10 pt, conference]{ieeeconf}  % Comment this line out if you need a4paper
\IEEEoverridecommandlockouts                                         % This command is only needed if you want to use the \thanks command
\usepackage[utf8]{inputenc}
\usepackage{array}
\usepackage{graphicx}
\usepackage{amsfonts}
\usepackage{amsmath}

\makeatletter
\let\NAT@parse\undefined

\newcommand{\Rmnum}[1]{\expandafter\@slowromancap\romannumeral #1@}
\makeatother
\usepackage[colorlinks, linkcolor=red, anchorcolor=blue, citecolor=green, bookmarks=false]{hyperref}

\title
{\LARGE \bf Vehicle Pose and Shape Estimation through Multiple Monocular Vision}

\author{Wenhao Ding$^{1}$, Shuaijun Li$^{2}$, Guilin Zhang$^{2}$, Xiangyu Lei$^{2}$ and Huihuan Qian$^{2}$% <-this % stops a space
\thanks{This research is supported by the NSFC project U1613226 from the State Joint Engineering Lab and Shenzhen Engineering Lab on Robotics and Intelligent Manufacturing, China.}% <-this % stops a space
\thanks{$^{1}$Wenhao Ding is with the Department of Electronic Engineering, Tsinghua University, Haidian District, Beijing, China  {\tt\small dingwenhao95@gmail.com}}%
\thanks{$^{2}$Shuaijun Li, Guilin Zhang, Xiangyu Lei and Huihuan Qian are with the Robotics and Artificial Intelligence Laboratory, The Chinese University of Hong Kong, Shenzhen, Shenzhen, China  {\tt\small sjli01@mae.cuhk.edu.hk, \{gl.zhang.cuhk, x.y.lei.hk\}@gmail.com, hhqian@cuhk.edu.cn}%
}
}

\begin{document}

\maketitle
\thispagestyle{empty}
\pagestyle{empty}

%%%%%%%%
\begin{abstract}
%%%%%%%%

In this paper, we present a method to estimate a vehicle’s pose and shape from off-board multi-view images. These images are taken from monocular cameras with small overlaps. We utilize state-of-the-art \textit{Convolutional Neural Networks (CNNs)} to extract vehicles’ semantic keypoints and introduce a \textit{Cross Projection Optimization (CPO)} method to estimate the 3D pose. During the iterative CPO process, an adaptive shape adjustment method named \textit{Hierarchical Wireframe Constraint (HWC)} is implemented to estimate the shape. Our approach is evaluated under both simulated and real-world scenes for performance verification. It’s shown that our algorithm outperforms other existing monocular and stereo methods for vehicles’ pose and shape estimation. This approach provides a new and robust solution for off-board visual vehicle localization and tracking, which can be applied to massive surveillance camera networks for intelligent transportation. 

\end{abstract}

%%%%%%%%%%%%%%
\section{INTRODUCTION}
%%%%%%%%%%%%%%

Most recently, road scene understanding is well studied for improving the perception ability of intelligent transportation. Meanwhile, 3D pose estimation for objects becomes a hot research topic, owing to its significance to the field of computer vision and robotics. These factors inspire us to focus on pose and shape estimation of vehicles to improve the perception ability of intelligent transportation.

Despite sensors like LiDAR, depth camera and stereo camera have been used for a long time, their application scopes are constrained due to high cost and other limitations. Therefore, to estimate vehicle’s pose information, more and more works concentrate on monocular visual estimation methods. These methods have potential to be applied to the massive surveillance camera network in real world.

In fact, mobile robots usually conduct on-board methods called simultaneous localization and mapping (SLAM) \cite{c10}. In contrast, off-board visual methods can also be considered for 3D pose estimation tasks. Considering that off-board methods have the advantage of possessing a better Field-of-View (FoV), and that many of the latest deep-learning based technics are developed based on off-board vision, there is a huge potential to use them for vehicle 3D pose estimation. 

\begin{figure}[h]
\centering
\includegraphics[width=7cm]{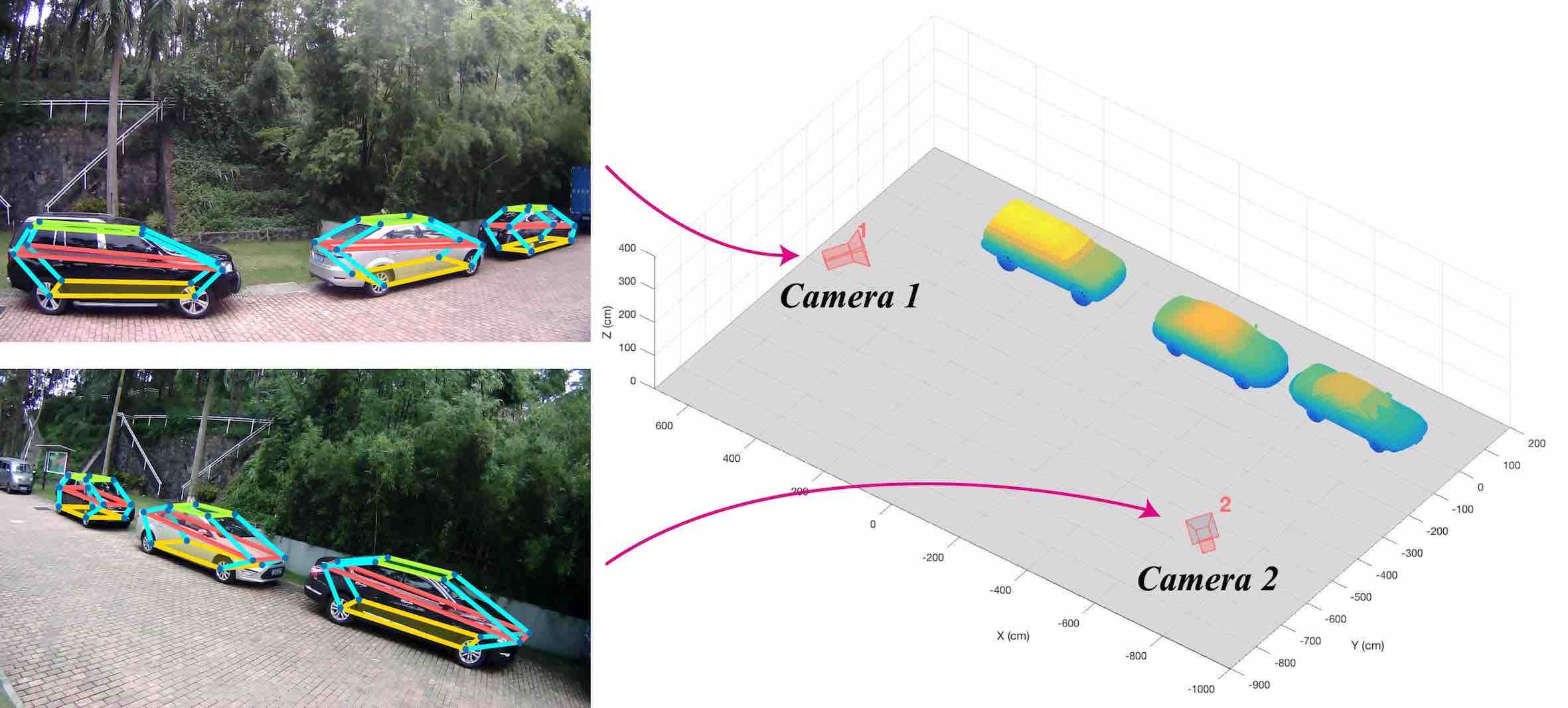}
\caption{\textbf{Example of pose and shape estimation.} 
Left two images come from two different cameras with small overlap. Wireframes projected on the images represent estimation results. On the right side, three vehicles are placed in a 3D space according to the estimated pose and CAD models.}
\end{figure}

For recent vision-based 3D pose estimation research, deep learning tools are widely used. Keypoints of objects are defined and detected to aid with the pose estimation \cite{c22}. Those CNN methods provide a new way to solve this kind of problem. But in most cases, single image is processed for tackling estimation task and this suffers from several drawbacks. As a remedy, it can be greatly improved if multiple images from different views can be used together in the scene of traffic monitoring. 

Based on above background, we propose an approach using multiple off-board cameras (two at least) with small overlaps to obtain 3D pose and shape of vehicles. An example of our approach is shown in Fig.\,1. In comparison to methods with bounding box annotation \cite{c4}, our approach utilizes a wireframe model to describe vehicle’s 3D pose and shape information. 

The whole algorithm is divided into two stages as shown in Fig.\,2. First, multiple images taken from cameras of different views (two images for simplification) are fed into a coarse-to-fine CNN that is trained for vehicle semantic keypoints detection specifically. After CNN processing, two sets of keypoints are obtained as outputs. Second, in optimization stage, CPO method projects a general 3D vehicle model onto each image with camera intrinsic and extrinsic parameters, and iteratively minimizes projection errors. It’s worth mentioning that no prior knowledge about the target vehicle is required in this approach. Vehicle’s shape estimation starts from a general wireframe model and is adjusted with HWC method.

%Using vehicle images from a robot simulation software environment and real world respectively (KITTI \cite{c33} is not suitable for traffic monitor application here because it doesn't have small-overlap image data), a series of tests have been conducted to evaluate the performance of our proposed approach. The evaluation results prove the accuracy and robustness of our estimating approach which has achieved less than $3^{\circ}$ of rotation error and $5\, cm$ of translation error. In pace with traffic surveillance's popularity in cities and rural areas, it is practical for those overlapping cameras to form a navigation system to satisfy automatic driving localization requirements, guide automatic parking system for collision avoidance, and realize other localization applications.

\begin{figure}[]
\centering
\includegraphics[width=8cm]{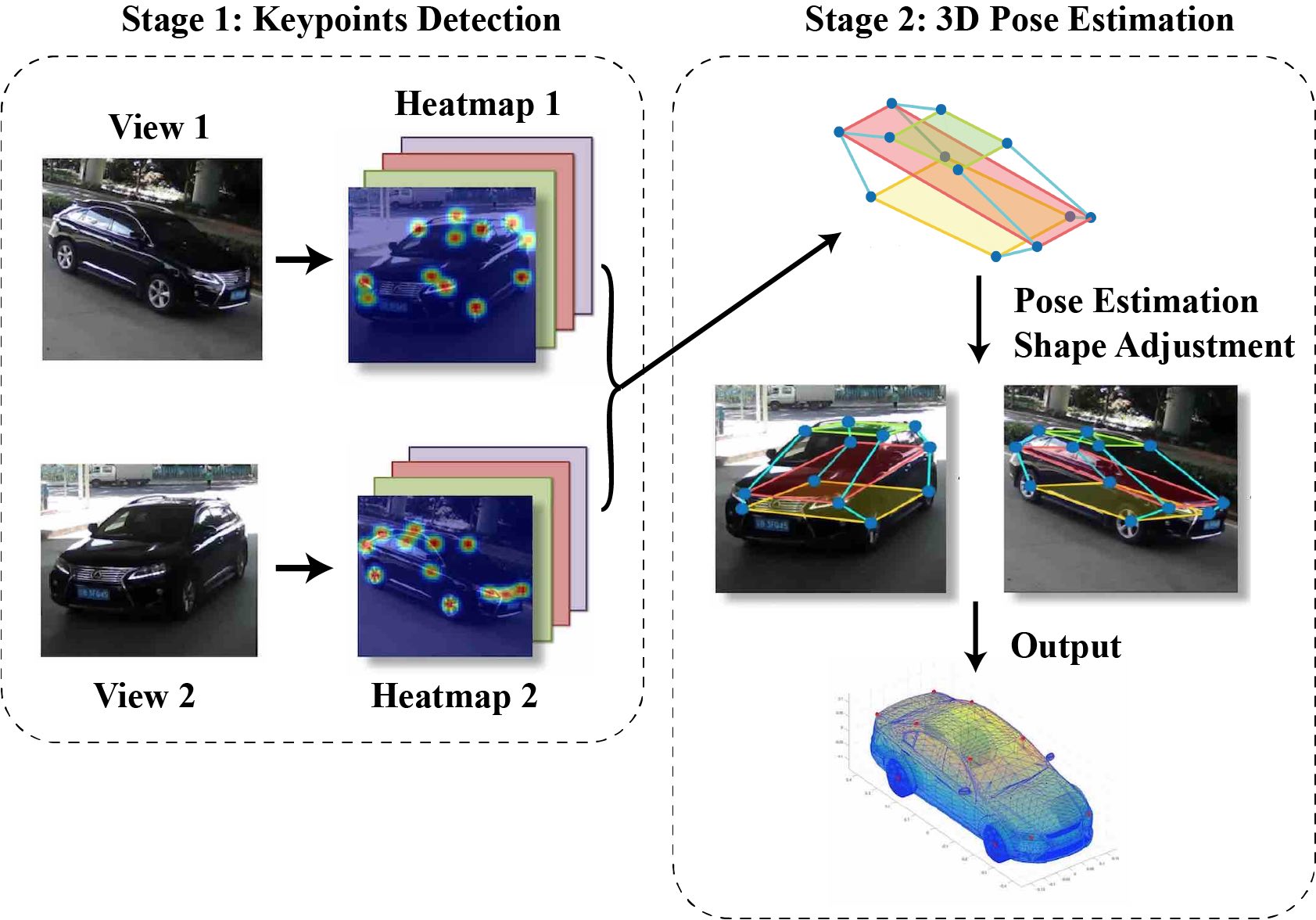}
\caption{\textbf{Overview of our approach.} 
Stage 1 consists the steps of CNN process which outputs heatmaps with highlight keypoints. Stage 2 utilizes results of the previous stage as input and start to fuse the information from multiple images. Our core algorithm comes next, with the methods of pose estimation (CPO) and shape adjustment (HWC), an accurate pose and shape estimation is obtained. Here, a CAD vehicle model is used for better display.  }
\end{figure}

%%%%%%%%%%%%%%%
\section{RELATED WORKS}
%%%%%%%%%%%%%%%

% 3D dataset is not easy to get
\subsection{Convolutional Neural Network}

The past few decades have witnessed the development of neural networks, especially in the field of complex feature detection. \cite{c2, c3} are state-of-the-art works of object detection. Ren \emph{et al.} \cite{c2} provided a real-time region proposal network to detect multiple objects on the 2D image. Joseph \emph{et al.} \cite{c3} proposed a detection method which can achieve the classification of 9,000 objects. All these detection methods help localization algorithm to focus mainly on their target.

For most detection and estimation tasks, CNN is used to recognize complex and high-level features, which cannot be sufficiently tackled by conventional vision methods. Lately, there are already some end-to-end methods for pose estimation. \cite{c12, c4} utilized video or multi-channel information obtained from on-board device to localize vehicles. \cite{c13} and \cite{c11} directly trained a CNN with a single image and 3D landmarks, but complex and large 3D datasets require much time and human resource for annotating. 

In the recent study, a stacked hourglass framework \cite{c1} is proposed to detect semantic keypoints on the bodies of human beings. Owing to its coarse-to-fine architecture, features can be detected on multiple resolutions, leading to high accuracy results. Compared to traditional random forests method like \cite{c16}, CNN outputs more accurate keypoints. Our approach follows this direction and takes advantages of these works.% We trained our CNN on a subset of PASCAL3D+ \cite{c30} and a simulation dataset. 

% single image method has defects
\subsection{Single Image Vehicle Localization}

As for robotics and intelligent transportation, vehicle localization is always at the cutting edge \cite{c23}. Both on-board and off-board methods have representative works.

Most of the off-board works combine a single image with complex vehicle models. Zhu \emph{et al.} \cite{c31} used constrained discriminative parts and pre-defined wireframe models to estimate the 3D pose. \cite{c19} proposed a similarity measure method with off-board camera, and a top-down perception approach is proposed in \cite{c32}. Works mentioned above all require complex accurate car models, which is usually unavailable. Besides, their feature detecting methods are not robust enough for arbitrary scene and viewpoint.

Some latest works \cite{c6, c7} combined semantic keypoints with simple models. They achieve large advantages in pose estimation task with the help of CNN. These approaches outperform most existing methods. However, methods with single image have a problem with translation error and model scale. Even if \cite{c6, c7} have shape adjustment in their process, this problem remains unsolved. According to the principle of 3D projection, depth of object depends on the scale of the model. Whether accurate pose is provided or not, the translation error increases when disproportionate model is used. Fig.\,3. shows a simple demonstrative experiment. We choose a wireframe which is smaller than groundtruth, and the left two columns show the results with only one camera. Disproportionate model generates right orientation but wrong translation and shape. As a contrast, the last column shows that two projects are both right. 

Besides the defect mentioned above,  the robustness of the keypoints detection is poor when vehicle occultation (by trees, walls, or other vehicles) occurs. To some extent, multiple camera approach can help with these limits.

\begin{figure}[h]
\centering
\includegraphics[width=8cm]{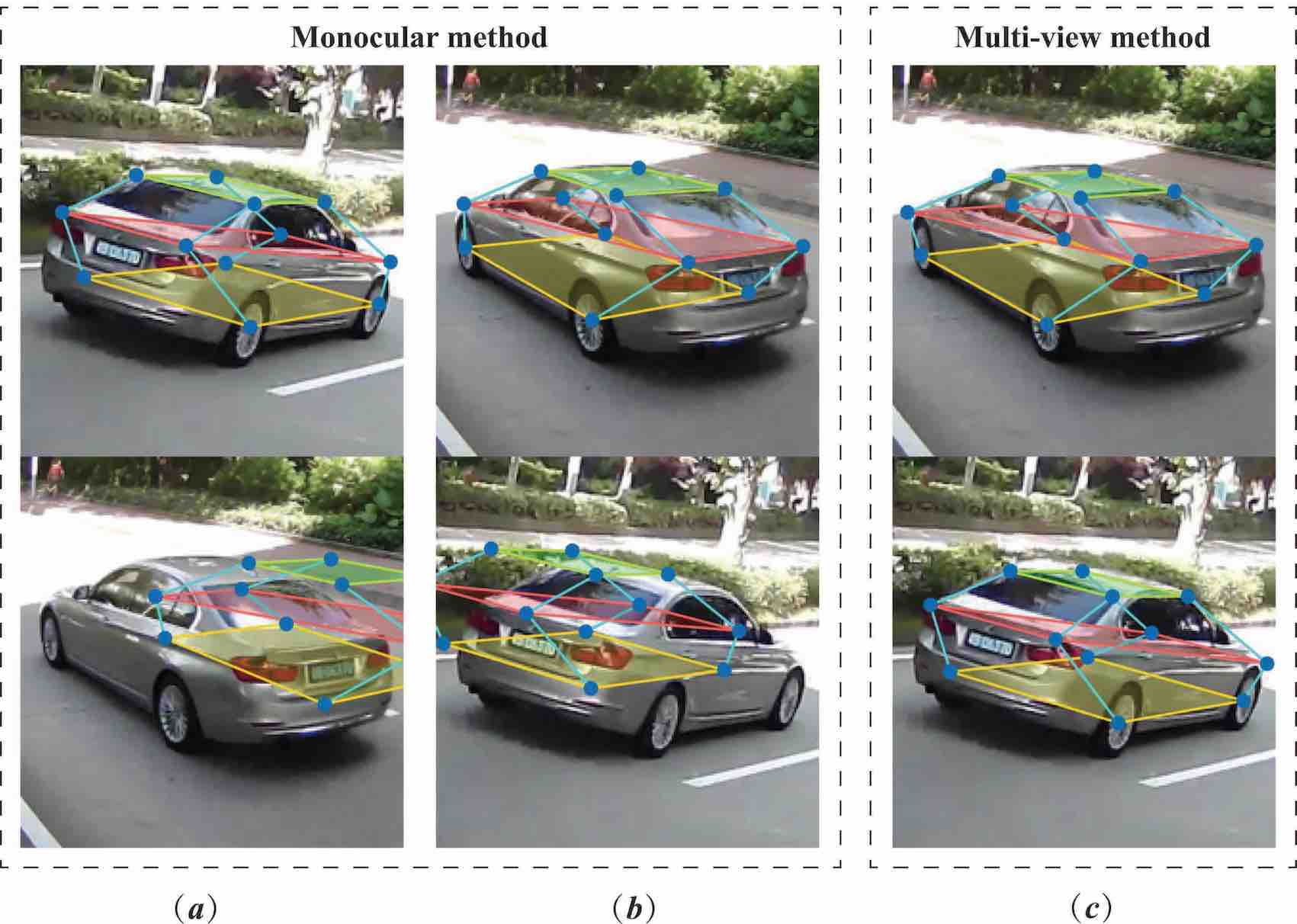}
\caption{\textbf{Comparison between monocular method and multi-view method.} 
(a) shows the result that only relies on camera 1, and the bottom image is the projection to another camera’s image plane. (b) is the same with (a), except for relying on camera 2. (c) are projection results achieved by CPO and HWC methods.}
\end{figure}

\begin{figure*}[]
\centering
\includegraphics[width=14cm]{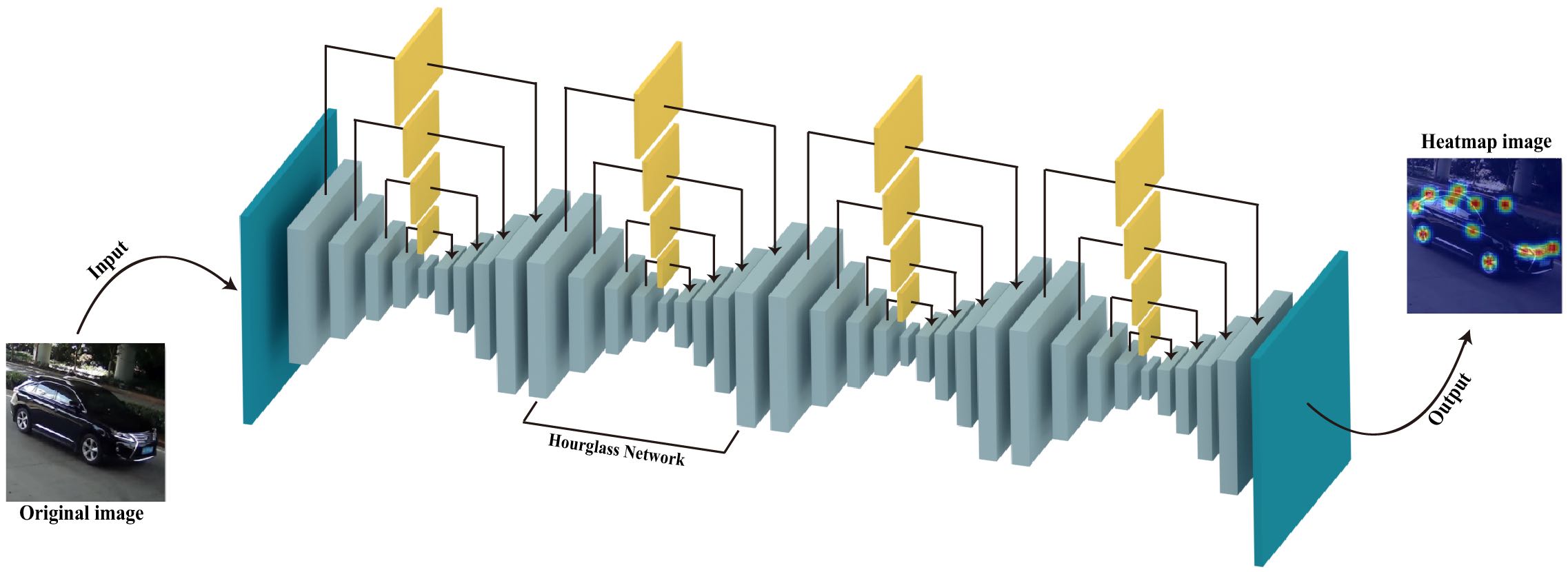}
\caption{\textbf{CNN with stacked hourglass architecture.} 
Four-layer hourglass architecture is used in our algorithm. Each sub-network has a intermedia output, which can be used for intermedia supervision. The loss function of is defined by the distance between the position of keypoints in label and output of CNN.}
\end{figure*} 

% stereo methods has defects
\subsection{Stereoscopic and Multiple Camera System}

The most similar algorithm with ours is stereoscopic algorithm. Stereo cameras usually consist two parallel cameras with small baseline or large overlap. Suppose using stereoscopic algorithm in our scenario. After getting two sets of keypoints of two images, we can calculate the 3D position of every keypoints with intrinsic and extrinsic parameters of camera. Then, we can connect those keypoints into a wireframe. But stereoscopic algorithm requires high precision, which means errors of keypoint from CNN lead to a large shift in 3D space. Still further, the shape of the vehicle will be asymmetrical. More results will be displayed and discussed at the experiment part.

Most multi-camera systems are applied to the field of human pose estimation, such as \cite{c14, c15, c16}. Pavlakos \emph{et al.} \cite{c16} utilized random forest to classify each pixel in each image, \cite{c14} recovered a volumetric prediction from multiple images. The standard principle of human pose estimation is using 3D pictorial structure, which has been proved to be effective. Our method is inspired by these approaches to combine image features with a deformable model.

To the best of our knowledge, there are few works of vehicle localization using multiple cameras. Chen \emph{et al.} \cite{c4} used bird view LiDAR data and front image to localize surrounding vehicles for the purpose of automatic driving, but only rough 3D bounding box results are given in this approach. \cite{c17} and \cite{c18} improved SLAM algorithm by integrating multi-view cameras and shape information. They outperformed some traditional SLAM methods. Calibrated stereo camera with small baseline can be used to reconstruct depth of the scene like \cite{c20} and \cite{c5}, but the narrow overlapping region is not wide enough for accurate vehicle localization. Although some defects still exist, works mentioned above prove the superiority of multi-view approaches and guide us to explore more possibilities in this direction.

% attack three points above
\subsection{Our Contribution}

Considering all the pros and cons of existing single-image and multi-image methods, we propose a new framework for vehicle 3D localization. Our contributions are as follow:

\begin{itemize}
\item We take advantage of hourglass architecture CNN to extract 2D feature from 2D annotation dataset. Training data and training process are much easier to implement.
\item Our methods can be applied to small overlap condition, which is more consistent with traffic monitoring camera system. 
\item Our multi-camera method has better performance over mono-camera method in aspects of precision and robustness of sheltered environments.
\item Inspired by deformable methods, we propose HWC to adaptively estimate vehicle's shape.
\end{itemize}

%%%%%%%%%%%%%%%%%
\section{METHOD}
%%%%%%%%%%%%%%%%%

The pipeline of our approach consists of two stages, keypoints detection and pose estimation. The process of pose estimation includes adaptive shape adjustment, with which we can describe the shape more accurate and improve the accuracy of estimation result.

\subsection{Semantic Keypoints Detection} 

Before keypoints detection, the Region of Interest (ROI) is required for selecting vehicles from images with 2D bounding box. As 2D object detection has been well studied for a period of time, we assume that ROIs have been provided by a state-of-the-art method \cite{c3} mentioned in Section \Rmnum{2}. 

Leveraging the recent success in keypoints detection like \cite{c23, c25}, we build and modify a four-layer hourglass network introduced by \cite{c1}, which is shown in Fig.\,4. This network framework has a symmetric architecture with residual module as its basic unit. This module contains both original information and high-level feature. One single hourglass network consists of several residual modules, and this hourglass sub-network achieves top-down and bottom-up process. In each hourglass, keypoint features can be extracted from both global and local resolutions. Another novel method used in this network is intermediate supervision. For every hourglass architecture, intermediate stage outputs a heatmap result, which can help solve the vanishing gradient problem. 

This architecture has gained great success in human body semantic keypoints localization and it is demonstrated that stacked architecture outperforms monolithic top-down networks \cite{c26, c27}. As for our task, this framework is trained on about 10,000 images from vehicle part of PASCAL3D+ as well as our own dataset. We find that a four-layer framework works well for our task. %Some detection results are sampled and shown in Fig.\,5.

%\begin{figure}[h]
%\centering
%\includegraphics[width=8.5cm]{f4.jpg}
%\caption{\textbf{Examples of our CNN output.} 
%Different views are chosen to show the robustness of this CNN framework, and all these images come from a multi-view dataset \cite{c28}. As result shows, blocked annotations can still be localized with relatively high confidence.}
%\end{figure} 

During the training stage, we input a label with 12 annotations (wheel$\times$4, light$\times$4, windshield$\times$2, rear window$\times$2) and the loss function are defined by the distance between label and CNN's output. It is 
\begin{equation}
\begin{aligned}
loss = \sum_{k=1}^{K}{||P_{label}-P_{output}||}^2
\end{aligned}
\end{equation}
where $P_{label}$ and $P_{output}$ represent the 2D position of 12 keypoints in label and CNN's output respectively. \textit{k} represents the index of keypoints and we use Euclidean distance to measure the difference.

A noteworthy thing is that self-concealed keypoints are also annotated, thus the network learns all keypoints of vehicle at the same time. Therefore, outputs of our network are 12 heatmaps, each of which has a 2D Gaussian distribution. The value of the Gaussian distribution denotes the probability of the keypoint position. Generally, obscured parts are always inaccurate in output heatmaps, and different views affect the CNN output as well. Then, in some cases, a single image may reason about wrong pose estimation with some keypoints that seriously deviate from real values. Considering this phenomenon, multi-view images are combined to avoid overly relying on one inaccurate image.

\begin{figure}[h]
\centering
\includegraphics[width=8cm]{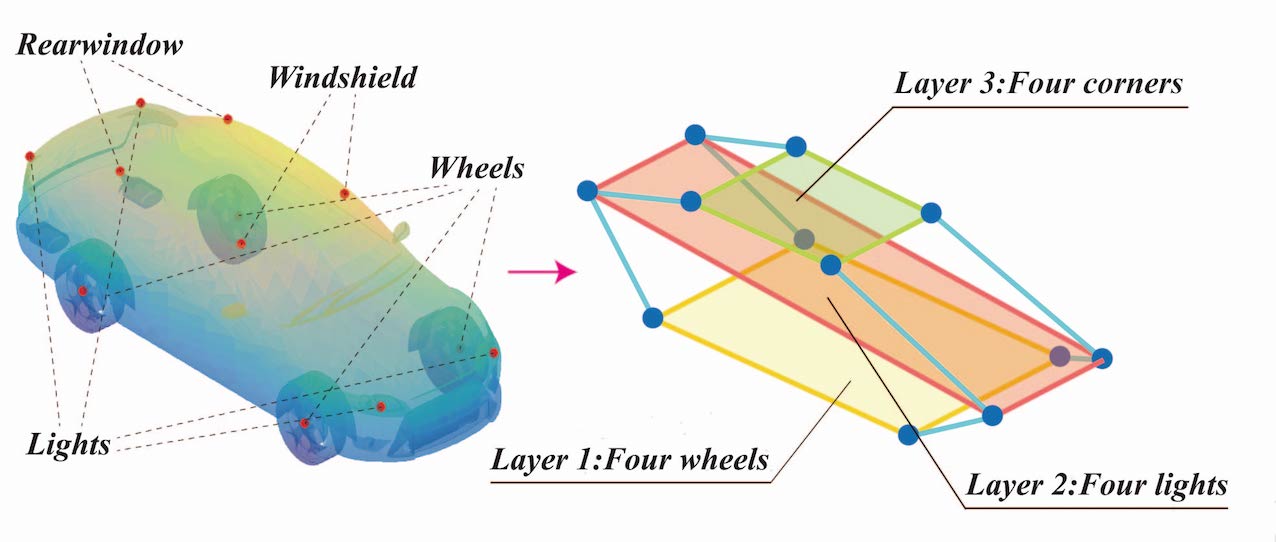}
\caption{\textbf{Explanation of HWC method.} 
A mean vehicle shape is acquired from several CAD models as the left image shows and four kinds of annotations are signed with notes. From this abstract model, a three-layer wireframe is introduced and constrained by some principles.}
\end{figure}

\begin{figure*}[]
\centering
\includegraphics[width=16cm]{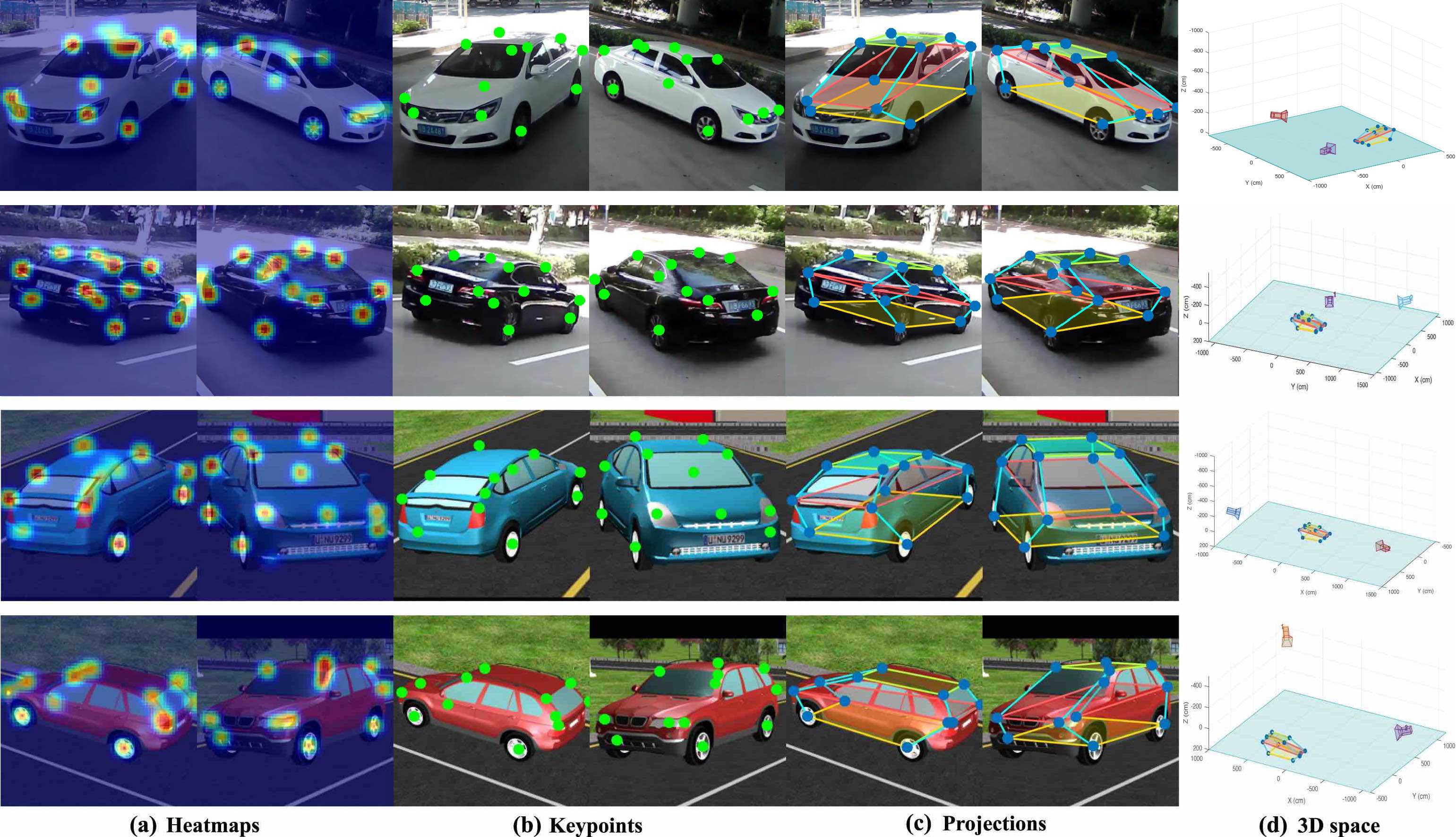}
\caption{\textbf{Qualitative results showing pose and shape estimation.} 
Every row displays a vehicle class and a different view combination. (a) shows the CNN outputs of two different views. (b) shows the keypoints of maximum probability in two cameras’ heatmaps. (c) shows projection of estimated pose and shape on both camera images. (d) shows results in 3D space.}
\end{figure*}

\subsection{Pose and Shape Estimation}

To accurately estimate 3D pose and shape of vehicles, we propose a method named CPO. This method projects a general 3D model to each image, and minimizes the projection error globally to get an accurate pose and shape. During the iterative process, the shape of the vehicle is adjusted under some constraints in HWC. For the sake of simplicity, we only consider the situation of two cameras.

\textbf{Initial Weight Matrix:} As we have semantic keypoints location from two images, we can integrate the information that which view has the best confidence for each keypoints. As two sets of heatmaps come from the same network, each value represents the confidence degree of the position of that keypoint, so subtracting the two point sets gives us the information about occultation and wrong detection. Then, a normalized weight can be obtained:
\begin{equation}
\begin{aligned}
\psi_{i, j}^k = &\mu_{1}w^{k}_{i, cnn} + \mu_{2} (w^{k}_{i, cnn}-w^{k}_{j, cnn})\\
&w^{k}_{i, nor} = \frac{\psi_{i, j}^k}{\sum_{k=1}^{K}\psi_{i, j}^k}
\end{aligned}
\end{equation}
The superscript $i$, $j$ means camera 1 and 2, $w^{k}_{i, cnn}$ represents the value of keypoint \textit{k} in camera 1. Then a diagonal weight matrix can be represented according to (2)
\begin{equation}
W = {\left[ \begin{array}{ccc}
w_{i, nor}^{i, 1} & \cdots & 0\\ 
\vdots& \ddots & \vdots\\ 
0& \cdots & w_{i, nor}^{i,K} 
\end{array} 
\right ]}
\end{equation}
Elements in this weight matrix are on behalf of the importance of each keypoint during the estimation process.

\textbf{Single Camera Iteration:} For each camera, we can estimate 3D pose from single image with a general model as approaches presented by \cite{c6} and \cite{c7}. Here we use singular value decomposition (SVD) method \cite{c29} for rigid motion estimation.

Denote the initial pose of our model by $\textit{P}$, and the number of keypoints is $\textit{K}$, then we can represent each 3D annotation as $\textit{p}_\textit{k}$ where $\textit{k}\, \in \, [\textit{1}, \textit{K}]$. The minimal projection error pose is denoted by $\textit{Q}$, thus $q_k$ represents the projection error pose of keypoint \textit{k}. Then we iterate a rotation $\mathcal{R}$ and a translation vector $\mathcal{T}$ such that 
\begin{equation}
(\mathcal{R}, \mathcal{T}) = \mathop{argmin}_{\mathcal{R}\in SO(d), \mathcal{T} \in \mathbb{R} ^d} \sum_{k=1}^{K} w_k \|(\mathcal{R}p_k+\mathcal{T})-q_k\|^2
\end{equation}
where $\textit{w}_\textit{k}$ means the weight getting from (2). Within several iterations, a pre-defined 3D model can be transformed close to the position with $\mathcal{R}$ and $\mathcal{T}$, where minimal projection error is obtained.  

\textbf{Multi-view Information Combination:} In one iteration, we do least-squares process for both cameras separately. After that, we minimize an energy function defined for cross projections:
\begin{equation}
f(t) = \frac{1}{2} \sum_{i=1}^{C} \left( \sum_{j=1}^{C} W(t)_i \|\pounds_{i, j}(t)\|^2 \right)
\end{equation}
where $\textit{t}$ denotes the iteration time and $\textit{\pounds}_{\textit{i}, \textit{j}}$ means the projection error from model optimized by camera $\textit{i}$ to image plane $\textit{j}$. \textit{C} represents the number of camera. Meanwhile, the weight matrix is updated along with iteration, and the criterion is defined for combining multi-view information:
\begin{equation}
\begin{aligned}
w_i^k (t + 1) = \mu_1 w_i^k (t) &+ \mu_2 \|\pounds_{i, j}(t)\|^2 \\&+ \mu_3 (w_{i, nor}^k (t) - w_{j, nor}^k (t))
\end{aligned}
\end{equation}
Here, the first term on the right-hand side means the weight value for keypoint $\textit{k}$ in last iteration, and the second term represents the projection error form camera $\textit{i}$ to image $\textit{j}$. The last item evaluates the visibility between two cameras. Updating weight matrix can help reaching a global maximum point for both cameras smoothly, as well as dynamically adjusting the importance of each keypoint.

\textbf{Hierarchical Wireframe Constraint:} With a view to vehicles, they all have a common general framework, in which every part has certain location. Moreover, inside the vehicle class, some degrees of freedom are available, such as the distance between light and wheel and the angle between wheel-plane and light-plane. Considering all the fixed and flexible rules, we design a constraint method for hierarchical models. Models under this principle are divided into three layers, as shown in Fig.\,6. These three layers are formed by 4 rooftop points, 4 light points and 4 wheel points, and each layer represents a plane. Since vehicles are highly symmetrical, we can define some general criterions during shape adjustment:

\begin{itemize}
\item Each layer should be vertical and symmetrical to the medial surface. 
\item Wheel layer should be a standard rectangle, while other two layers can be extended into a trapezoidal in a certain range.
\item The relationship between layers has a high degree of freedom for different kinds of vehicles. 
\item Distance between two points is flexible but within certain maximum. 
\end{itemize}

%\begin{figure*}[]
%\centering
%\includegraphics[width=16cm]{f7.jpg}
%\caption{\textbf{Qualitative results showing pose and shape estimation of real cars.} 
%Every row displays a vehicle class and a different view combination. (a) shows the CNN outputs of two different views. (b) shows the keypoints of maximum probability in two cameras’ heatmaps. (c) shows 	%projection of estimated pose and shape on both camera images. (d) shows a 3D space where real vehicle wireframe  models are placed in.}
%\end{figure*}

\begin{table*}[]
\setlength{\tabcolsep}{5pt}
\renewcommand\arraystretch{1.5}
\centering
\caption{translation errors of 12 keypoints. Three methods (Stereoscopic, 6DoF\cite{c6}, Ours) are compared.}
\begin{tabular}{c||c|c|c|c|c|c|c|c|c|c|c|c|c}
\hline
\textbf{Approaches} & \textbf{\begin{tabular}[c]{@{}c@{}}P1\end{tabular}} & \textbf{\begin{tabular}[c]{@{}c@{}}P2\end{tabular}} & \textbf{\begin{tabular}[c]{@{}c@{}}P3\end{tabular}} & \textbf{\begin{tabular}[c]{@{}c@{}}P4\end{tabular}} & \textbf{\begin{tabular}[c]{@{}c@{}}P5\end{tabular}} & \multicolumn{1}{l|}{\textbf{\begin{tabular}[c]{@{}l@{}}P6\end{tabular}}} & \textbf{\begin{tabular}[c]{@{}c@{}}P7\end{tabular}} & \textbf{\begin{tabular}[c]{@{}c@{}}P8\end{tabular}} & \textbf{\begin{tabular}[c]{@{}c@{}}P9\end{tabular}} & \textbf{\begin{tabular}[c]{@{}c@{}}P10\end{tabular}} & \multicolumn{1}{l|}{\textbf{\begin{tabular}[c]{@{}l@{}}P11\end{tabular}}} & \textbf{\begin{tabular}[c]{@{}c@{}}P12\end{tabular}} & \textbf{Mean} \\ \hline
{Stereoscopic-mean}         & 34.34    & 29.58    & 56.13     & 65.76    & 31.53     & 43.38   & 39.51     & 49.62  & 46.13    & 54.70   & 67.21    & 73.71   & 49.30        \\ \hline
{6DoF-LINCON {[}6{]}}      & 18.73    & 14.50     & 28.29     & 30.73     & 12.33     & 8.99    & 27.97     & 32.39  & 15.84   & 12.95   & 58.60    & 55.71    & 26.42           \\ 
{6DoF-TOYOTA {[}6{]}}      & 37.21   & 42.33     & 60.02     & 66.87     & 27.14     & 29.91   & 75.74    & 75.01  & 50.17    & 53.13   & 87.28    & 82.33    & 58.10         \\ 
{6DoF-BMW {[}6{]}}           &  29.65   & 28.60     & 40.09     & 39.39    & 23.98     & 25.80   & 45.84     & 40.84  & 15.44    & 10.95   & 25.75    & 23.05    & 29.12        \\ 
{6DoF-mean {[}6{]}}           & 28.53    & 26.48     & 46.14     & 45.66    & 21.15     & 21.57   & 49.85     & 49.41  & 27.15    & 25.68   & 57.21    & 53.70    & 37.88        \\ \hline
{Ours-LINCON}                 & 7.30      & 7.39       & 6.59       & 5.12      & 6.03       & 5.90     & 13.32     & 12.56  & 7.17      & 6.98     & 25.48    & 24.58    & 10.78        \\
{Ours-TOYOTA}                & 8.48      & 10.21     & 10.52     & 8.11      &  7.30      & 7.17     & 14.20    & 14.20    & 7.99    &  7.86     & 32.87    & 33.33     & 13.52     \\ 
{Ours-BMW}                      &{3.15}      &{2.94} &{3.79} &{4.60} &{4.96} &{4.93} &{8.99} &{9.96} &{5.96} &{4.73} &{17.41} &{14.03} &{7.12} \\ 
\textbf{Ours-mean}   & \textbf{6.31}&\textbf{6.85}&\textbf{6.97}&\textbf{5.94}&\textbf{6.10} &\textbf{6.01}&\textbf{12.17}&\textbf{12.24} &\textbf{7.04}&\textbf{6.52} &\textbf{25.25}&\textbf{23.98}&\textbf{10.48} \\ \hline
\end{tabular}
\end{table*}

\textbf{Shape Adjustment:} With the HWC method mentioned above, we can gradually correct the wireframe model during the pose estimating process. Some methods like \cite{c7} separated pose estimation and shape adjustment into two stages, but it is very possible that inaccurate shape leads pose estimation into local minimum points.

When doing shape adjustment, we optimize each point with (4) using constrained optimization algorithm for multivariate functions. Four points which are in the same layer are considered together. For each layer, HWC method is applied. In order to keep a symmetric architecture, a weighted average method is introduced:

\begin{equation}
\begin{aligned}
p^k (x, y) = &\,w_{i, project}^k p_i^k (x, y) + w_{j, project}^k p_j^k (x, y)\\
&\textbf{s.t.}  \,\,w_{i, project}^k + w_{j, project}^k = 1
\end{aligned}
\end{equation}

where $\textit{k}$ means the number of keypoints, and two normalized weight coefficients come from projection error and keypoint weight value. $\textit{p}^{\textit{k}}\textit{(x,\,y)}$ denotes the new 3D position of keypoint $\textit{k}$. Across layers, maximal moving step size is set in case of resulting in a large deformation. Doing shape adjustment with totally wrong pose is risky, so some pure pose estimation iterations are processed firstly until a relatively smooth energy function value is obtained.
After reaching the maximum iteration or minimum threshold between two adjacent iterations, the whole process stops. In most instances, optimized results from two cameras are almost the same, but when there is a deviation between them, we choose the smaller projection error one as our final result.

%%%%%%%%%%%%%
\section{EXPERIMENTS}
%%%%%%%%%%%%%

In this section, we evaluate the performance of our algorithm under simulated and real-world conditions (KITTI \cite{c33} is not suitable for traffic monitor application here because it doesn't have small-overlap image data). Qualitative results of the test under two conditions are shown in Fig.\,7. 

Webots is a development environment used to model and simulate mobile robots. Through building up simulation environment and importing vehicle models from the database, our algorithms can be quickly and effectively verified. Besides, accurate ground truths of models can be accessed easily according to the coordinate system inside the software. The extrinsic and intrinsic parameters of cameras used in the experiments can also be calibrated of convenience with a chessboard. Moreover, real-world experiments are also conducted to ensure the practicality and robustness of our algorithm. Test results of a series of experiments using real vehicles are covered in this section.

\subsection{Simulated Experiments using Webots}

We use Webots as our simulation platform to collect data for CNN and our proposed algorithm. Three kinds of vehicle models were imported from 3D database, including Lincoln MKZ, Toyota Prius and BMW X5. We call two images with known camera parameters one group. For each model, we tested more than 50 groups. Objects in Webots are observed with perspective projection. Then, a chessboard with size of $2m \times 3m$ was chosen to calibrate the extrinsic and intrinsic parameters of individual monocular cameras (less than 0.2 pixels error). Ground truths represented by 3D coordinates of the 12 semantic keypoints were recorded.  

Images taken from different views were used to predict the semantic keypoints of vehicles using hourglass CNN. Heatmaps were produced to record the gaussian distribution of the predicted semantic keypoints. To evaluate our approach, we consider errors of this experiment in two different metrics. First, the errors of the 12 semantic keypoints of vehicles with respect to those of the groundtruth, which can be expressed as euclidean distance (cm). Second, the errors of orientation and translation of the estimated pose with respect to the ground truths can be expressed as below:
\begin{equation}
\Delta(R_1, R_2) = \frac{\|log (R_1^{T}R_2)\|_F}{\sqrt{2}}
\end{equation}
\begin{equation}
\Delta(T_1, T_2) = \|\frac{\sum_{k=1}^{K}p^k_1}{K}-\frac{\sum_{k=1}^{K}p^k_2}{K}\|
\end{equation}
where the rotation error is represented as geodesic distance and the translation error is represented as distance between the centroids of two point sets. 

We use stereoscopic method as baseline in our experiments. After obtaining camera parameters and heatmaps from CNN, stereoscopic formula can be used to calculate 3D locations of each keypoint.
%\begin{equation}
%\left\{
%\begin{aligned}
%z & = \frac{f_l(f_rt_y-Y_rt_z)}{Y_r(r_7X_l+r_8Y_l+f_lr_9)-f_r(r_4X_l+r_5Y_l+f_lr_6)} \\
%x & =  \frac{zX_l}{f_l} , y =  \frac{zY_l}{f_l} 
%\end{aligned}
%\right.
%\end{equation}

One thing to note is that most localization and pose estimating tasks are trained and evaluated on KITTI \cite{c33}, so they usually use mean Average Precision (mAP) and Average Orientation Similarity (AOS) as metric criterions. But in the application of traffic monitor, on-board stereo images are not applicable. Considering some works using rotation and translation error as criterion, we decide to follow them to make the contrast more intuitive.

Keypoints localization errors in 3D space are shown in Table \Rmnum{1}, including comparisons with stereoscopic baseline a state-of-the-art method \cite{c6}. The order of the points is Left Front Wheel, Right Front Wheel, Left Back Wheel, Right Back Wheel, Left Front Light, Right Front Light, Left Back Light, Right Back Light, Left-up Windshield, Right-up Windshield, Left-up Rear Window, Right-up Rear Window. We notice that, due to the HWC shape adjustment method, although sometimes CNN predictions are not accurate, final keypoint positions are extremely close to the groundtruth with our approach. \cite{c6} uses only single image, so we recorded the best result of two images. After testing algorithms on three different kinds of vehicle (each has more than 100 images), we reach the conclusion that less keypoint distance error is achieved compared with single image approach \cite{c6}. It is worth discussing that rear window points and tail lamp points are less accurate than others. One reason to explain is those keypoints change relatively large inside vehicle classes, resulting in inaccurate output of CNN keypoints detector. Despite with these detection errors, rotation and translation error can still be controlled below a low level. 
 
\begin{table}[h]
	\renewcommand\arraystretch{1.5}
	\centering
	\caption{Rotation and  translation error of the whole vehicle}
	\label{my-label}
	\begin{tabular}{c||c|c|c||c}
		\hline
		\textbf{Approaches}        & \multicolumn{3}{c||}{\textbf{Rotation (degree)}} & \textbf{Translation (cm)} \\ \hline
		{Stereoscopic Method}   & \multicolumn{3}{c||}{10.56}                      & 44.32                   \\ \hline
		{PNP}                              & \multicolumn{3}{c||}{7.17}                       & 36.64                   \\ \hline
		{6DoF-weak {[}6{]}}         & \multicolumn{3}{c||}{7.99}                       & N/A                       \\ \hline
		{6DoF-full {[}6{]}}             & \multicolumn{3}{c||}{5.57}                       & 27.57                   \\ \hline
		{Viewpoint {[}23{]}}          & \multicolumn{3}{c||}{9.10}                       & N/A                       \\ \hline
		{Reconstruct {[}7{]}}        & 8.79           & 12.57          & 16.16          & N/A                       \\ \hline
		{ObjProp3D {[}5{]}}          & 17.37         & 21.86         & 26.87         & N/A                       \\ \hline
		{3DVP {[}13{]}}                & \multicolumn{3}{c||}{11.18}                    & N/A                       \\ \hline
		{Ours-Webots}                  & \multicolumn{3}{c||}{4.4134}                  & 6.21                    \\ \hline
		\textbf{Ours-Real World}             & \multicolumn{3}{c||}{\textbf{2.87}}       & \textbf{4.73}             \\ \hline
	\end{tabular}
\end{table}   

\subsection{Real World Experiments}

This section verifies the performances of our algorithm under real-world conditions. The pipeline of conducting real-world experiments is similar to that of the simulation experiments. A series of vehicles representing different car models were selected to be processed by the proposed algorithm. Groundtruth was recorded, and the cameras were calibrated with a chess board in advance.

Similarly, we consider errors of the estimated poses and translation with (7) and (8). Results are recorded in Table \Rmnum{2}, including comparisons with other state-of-the-art algorithms. The baseline algorithm is stereoscopic method. One thing needs to be claimed is translation error cannot be obtained for some monocular methods. For \cite{c7} and ObjProp3D \cite{c5}, those three results are obtained on three different difficulty level datasets of KITTI \cite{c33}.

%At last, we discuss the defect of single-image method in detail. When the recovered 3D models of vehicles generated by the single monocular image algorithm is projected to the imaging plane of another camera, the estimated pose is usually observed to be inaccurate. The explanation for the occurrence of these results is that the scale of the deformable model is not determined accurately with selected vehicle models, and then, the depth from the recovered model to the imaging plane is not accurately determined. The situation is shown in Fig. 8, where the original model is smaller than real size.

%%%%%%%%%%%%%%
\section{CONCLUSIONS}
%%%%%%%%%%%%%%

We propose an accurate approach to estimate pose and shape of vehicles in this paper. A cross projection optimization scheme and a deformable model constrain method are implemented to make the best of the information from multiple images.  We evaluate our method on both simulated platform and real world, and demonstrate superior performance than published monolithic and stereo algorithms. We achieved less than $3^\circ$ and $5\,cm$ error in aspect of rotation and translation.

Even with imprecise keypoint localization, our method still presents robustness. Moreover, owing to the accurate wireframe description of the shape, precise 3D dimensions can improve the ability of collision avoidance and motion planning in transportation and mobile robot. 

In terms of application prospects of our approach, modern surveillance camera networks have enough resolution for object detection and pose estimation. Additionally, surveillance cameras usually have common visual field which gives our algorithm an opportunity to be applied. Based on these hardware systems, more accurate vehicle localization can be achieved with the help of our algorithm.

%%%%%%%%%%%%%%

\end{document}